\documentclass{sig-alternate-05-2015}

\usepackage[ruled]{algorithm2e}
\usepackage{tabularx}
\usepackage{caption}
\usepackage{subfigure}

\begin{document}


\clubpenalty=10000
\widowpenalty = 10000

\title{UnitBox: An Advanced Object Detection Network}

\numberofauthors{1} 
%
\author{
%
%
\alignauthor{
Jiahui Yu$^{1,2}$\ \ \ \ \
Yuning Jiang$^2$ \ \ \ \ \
Zhangyang Wang$^1$\ \ \ \ \
Zhimin Cao$^2$\ \ \ \ \
Thomas Huang$^1$\\
}
\affaddr{
\vspace{0.2cm}
$^1$University of Illinois at Urbana$-$Champaign
\\}
\affaddr{
$^2$Megvii Inc\\
}
\affaddr{\{jyu79, zwang119, t-huang1\}@illinois.edu, \{jyn, czm\}@megvii.com}
}

\maketitle
\begin{abstract}
In present object detection systems, the deep convolutional neural networks
(CNNs) are utilized to predict bounding boxes of object candidates, and have
gained performance advantages over the traditional region proposal methods.
However, existing deep CNN methods assume the object bounds to be four
independent variables, which could be regressed by the $\ell_2$ loss
separately. Such an oversimplified assumption is contrary to the well-received
observation, that those variables are correlated, resulting to less accurate
localization. To address the issue, we firstly introduce a novel Intersection
over Union ($IoU$) loss function for bounding box prediction, which regresses
the four bounds of a predicted box as a whole unit. By taking the advantages of
$IoU$ loss and deep fully convolutional networks, the UnitBox is introduced,
which performs accurate and efficient localization, shows robust to objects of
varied shapes and scales, and converges fast. We apply UnitBox on face
detection task and achieve the best performance among all published methods on
the FDDB benchmark.

\end{abstract}

\keywords{Object Detection; Bounding Box Prediction; $IoU$ Loss}

\section{Introduction}

Visual object detection could be viewed as the combination of two tasks: object
localization (where the object is) and visual recognition (what the object
looks like). While the deep convolutional neural networks (CNNs) has witnessed
major breakthroughs in visual object recognition~\cite{cv:ResNet} \cite{cv:VGG}
\cite{cv:zhangyang}, the CNN-based object detectors have also achieved the
state-of-the-arts results on a wide range of applications, such as face
detection~\cite{cv:deep_cascade} \cite{cv:DenseBox}, pedestrian
detection~\cite{cv:swpshui} \cite{cv:DeepP} and etc~\cite{cv:RCNN}
\cite{cv:xinchao} \cite{cv:FasterRCNN}.

Currently, most of the CNN-based object detection
methods~\cite{cv:RCNN} \cite{cv:DeepP} \cite{cv:deep_cascade} could be summarized
as a three-step pipeline: firstly, region proposals are extracted as object
candidates from a given image. The popular region proposal methods include
Selective Search~\cite{cv:SS}, EdgeBoxes~\cite{cv:EdgeBoxes}, or the early
stages of cascade detectors~\cite{cv:deep_cascade}; secondly, the extracted
proposals are fed into a deep CNN for recognition and categorization; finally,
the bounding box regression technique is employed to refine the coarse proposals
into more accurate object bounds. In this pipeline, the region proposal
algorithm constitutes a major bottleneck in terms of localization effectiveness,
as well as efficiency. On one hand, with only low-level features, the
traditional region proposal algorithms are sensitive to the local appearance
changes, e.g., partial occlusion, where those algorithms are very likely to
fail. On the other hand, a majority of those methods are typically based on
image over-segmentation~\cite{cv:SS} or dense sliding
windows~\cite{cv:EdgeBoxes}, which are computationally expensive and have hamper
their deployments in the real-time detection systems.

\begin{figure}
 \centering
 \includegraphics[width=0.45\textwidth]{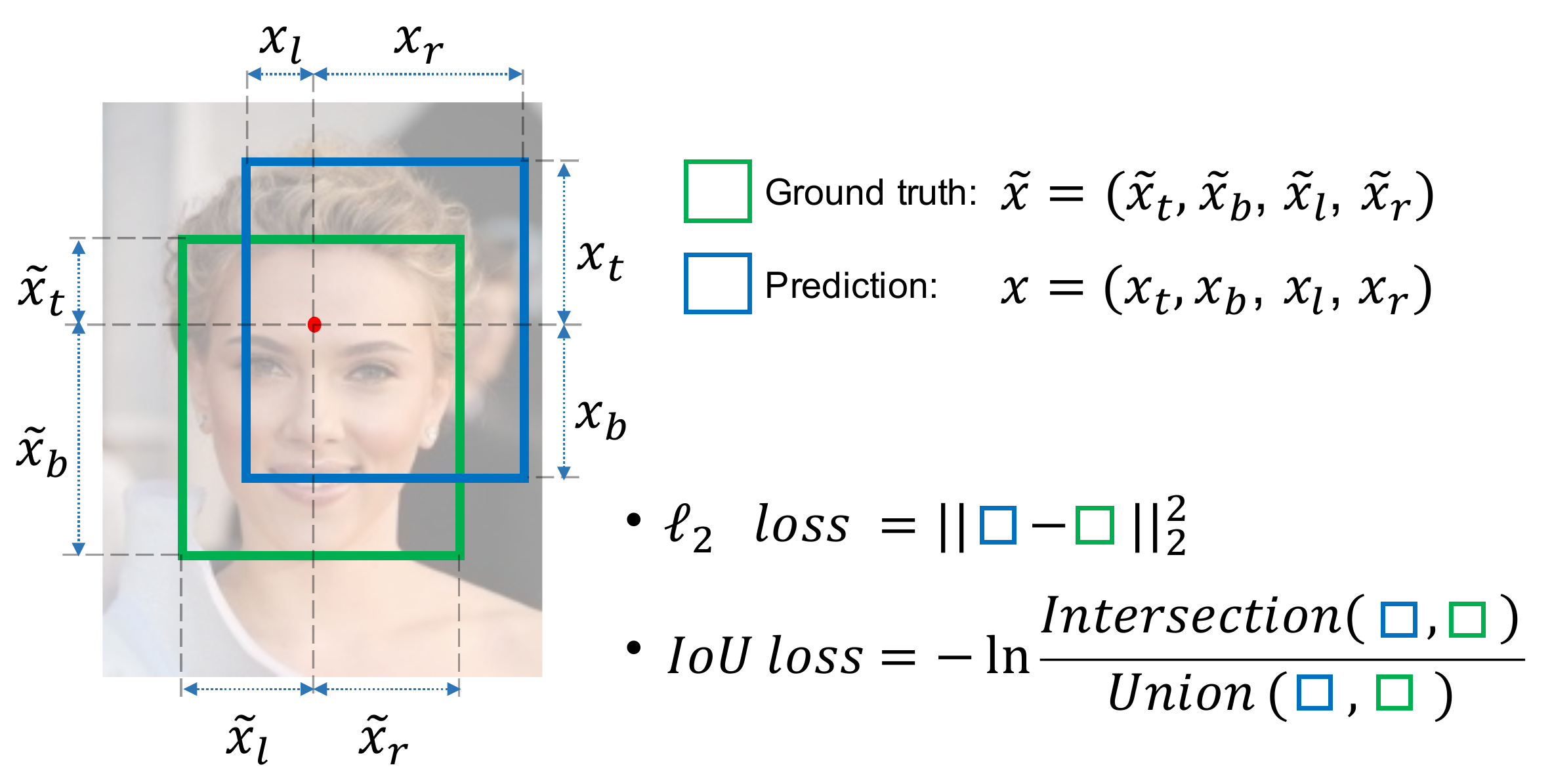}
 \caption{Illustration of $IoU$ loss and $\ell_2$ loss for pixel-wise bounding box prediction.}
 \label{fig:introduction}
\end{figure}

To overcome these disadvantages, more recently the deep CNNs are also applied to
generate object proposals. In the well-known Faster R-CNN
scheme~\cite{cv:FasterRCNN}, a region proposal network (RPN) is trained to
predict the bounding boxes of object candidates from the {\it anchor} boxes.
However, since the scales and aspect ratios of {\it anchor} boxes are
pre-designed and fixed, the RPN shows difficult to handle the object candidates
with large shape variations, especially for small objects.


Another successful detection framework, DenseBox~\cite{cv:DenseBox}, utilizes
every pixel of the feature map to regress a 4-D distance vector (the distances
between the current pixel and the four bounds of object candidate containing
it). However, DenseBox optimizes the four-side distances as four independent
variables, under the simplistic $\ell_2$ loss, as shown in
Figure~\ref{fig:introduction}. It goes against the intuition that those
variables are correlated and should be regressed jointly.

Besides, to balance the bounding boxes with varied scales, DenseBox requires the
training image patches to be resized to a fixed scale. As a consequence,
DenseBox has to perform detection on image pyramids, which unavoidably affects
the efficiency of the framework.


The paper proposes a highly effective and efficient CNN-based object detection
network, called UnitBox. It adopts a fully convolutional network architecture,
to predict the object bounds as well as the pixel-wise classification scores on
the feature maps directly. Particularly, UnitBox takes advantage of a novel
Intersection over Union ($IoU$) loss function for bounding box prediction. The
$IoU$ loss directly enforces the maximal overlap between the predicted bounding
box and the ground truth, and jointly regress all the bound variables as a whole
unit (see Figure~\ref{fig:introduction}). The UnitBox demonstrates not only more
accurate box prediction, but also faster training convergence. It is also
notable that thanks to the $IoU$ loss, UnitBox is enabled with variable-scale
training. It implies the capability to localize objects in arbitrary shapes and
scales, and to perform more efficient testing by just one pass on singe scale.
We apply UnitBox on face detection task, and achieve the best performance on
FDDB~\cite{data:FDDB} among all published methods.

\section{IoU Loss Layer} \label{sec:iou_loss}
Before introducing UnitBox, we firstly present the proposed $IoU$ loss layer and
compare it with the widely-used $\ell_2$ loss in this section.  Some important
denotations are claimed here: for each pixel $(i,j)$ in an image, the bounding
box of ground truth could be defined as a 4-dimensional vector:
\begin{equation}
 \boldsymbol{\widetilde x_{i,j}} =
 (\widetilde x_{t_{i,j}}, \widetilde x_{b_{i,j}}, \widetilde x_{l_{i,j}}, \widetilde x_{r_{i,j}}),
\end{equation}
where $\widetilde x_t$, $\widetilde x_b$, $\widetilde x_l$, $\widetilde x_r$
represent the distances between current pixel location $(i,j)$ and the top,
bottom, left and right bounds of ground truth, respectively.  For simplicity,
we omit footnote $i,j$ in the rest of this paper.  Accordingly, a predicted
bounding box is defined as $\boldsymbol{x} = (x_t, x_b, x_l, x_r)$, as shown in
Figure~\ref{fig:introduction}.

\subsection{L2 Loss Layer} \label{ssec:ssec2.1}
$\ell_2$ loss is widely used in optimization.
In~\cite{cv:DenseBox} \cite{cv:swshui}, $\ell_2$ loss is also employed to
regress the object bounding box via CNNs, which could be defined as:
\begin{equation}
 \mathcal{L}(x, \widetilde x) = \sum_{i \in \{t, b, l, r\}}{(x_i - \widetilde x_i)^2},
 \label{Eqn:l2}
\end{equation}
where $\mathcal{L}$ is the localization error.

However, there are two major drawbacks of $\ell_2$ loss for bounding box
prediction.  The first is that in the $\ell_2$ loss, the coordinates of a
bounding box (in the form of $x_t$, $x_b$, $x_l$, $x_r$) are optimized as four
independent variables.  This assumption violates the fact that the bounds of an
object are highly correlated.  It results in a number of failure cases in which
one or two bounds of a predicted box are very close to the ground truth but the
entire bounding box is unacceptable; furthermore, from Eqn.~\ref{Eqn:l2} we can
see that, given two pixels, one falls in a larger bounding box while the other
falls in a smaller one, the former will have a larger effect on the penalty
than the latter, since the $\ell_2$ loss is unnormalized.  This unbalance
results in that the CNNs focus more on larger objects while ignore smaller
ones. To handle this, in previous work~\cite{cv:DenseBox} the CNNs are fed with
the fixed-scale image patches in training phase, while applied on image
pyramids in testing phase.  In this way, the $\ell_2$ loss is normalized but
the detection efficiency is also affected negatively.

\subsection{IoU Loss Layer: Forward}
In the following, we present a new loss function, named the $IoU$ loss, which
perfectly addresses above drawbacks.  Given a predicted bounding box
$\boldsymbol x$ (after ReLU layer, we have $x_t, x_b, x_l, x_r \geq 0$) and the
corresponding ground truth $\boldsymbol{\widetilde x}$, we calculate the $IoU$
loss as follows:

\begin{algorithm}[h]
\SetAlgoNoLine
\textbf{Input: }{$\boldsymbol{\widetilde x}$ as bounding box ground truth} \\
\textbf{Input: }{$\boldsymbol{x}$ as bounding box prediction} \\
\textbf{Output: }{$\mathcal{L}$ as localization error} \\
\For{each pixel $(i,j)$} {
 \eIf{$\boldsymbol{\widetilde x} \neq \boldsymbol{0}$} {
  $X = (x_t + x_b) * (x_l + x_r)$\\
  $\widetilde X = (\widetilde x_t + \widetilde x_b) * (\widetilde x_l + \widetilde x_r)$\\
  $I_h = min(x_t, \widetilde x_t) + min(x_b, \widetilde x_b)$\\
  $I_w = min(x_l, \widetilde x_l) + min(x_r, \widetilde x_r)$\\
  $I = I_h * I_w$\\
  $U = X +\widetilde X - I$\\
  $IoU = \dfrac{I}{U}$\\
  $\mathcal{L} = -ln(IoU)$\\
  } {
   $\mathcal{L} = 0$
  }
 }
\caption{$IoU$ loss Forward}
\end{algorithm}

In Algorithm 1, $\boldsymbol{\widetilde x} \neq \boldsymbol{0}$ represents that
the pixel $(i,j)$ falls inside a valid object bounding box; $X$ is area of the
predicted box; $\widetilde X$ is area of the ground truth box; $I_h$, $I_w$ are
the height and width of the intersection area $I$, respectively, and $U$ is the
union area.

Note that with $0 \leq IoU \leq 1$, $\mathcal{L} = -ln(IoU)$ is essentially a
cross-entropy loss with input of $IoU$: we can view $IoU$ as a kind of random
variable sampled from Bernoulli distribution, with $p (IoU=1) = 1$, and the
cross-entropy loss of the variable $IoU$ is $\mathcal{L} = -pln(IoU) -
(1-p)ln(1-IoU) = -ln(IoU)$.  Compared to the $\ell_2$ loss, we can see that
instead of optimizing four coordinates independently, the $IoU$ loss considers
the bounding box as a unit. Thus the $IoU$ loss could provide more accurate
bounding box prediction than the $\ell_2$ loss. Moreover, the definition
naturally norms the $IoU$ to $[0,1]$ regardless of the scales of bounding
boxes. The advantage enables UnitBox to be trained with multi-scale objects and
tested only on single-scale image.

\subsection{IoU Loss Layer: Backward}
To deduce the backward algorithm of $IoU$ loss, firstly we need to compute the
partial derivative of $X$ w.r.t. $x$, marked as $\nabla _xX$ (for simplicity,
we notate $x$ for any of $x_t$, $x_b$, $x_l$, $x_r$ if missing):
\begin{equation}
  \frac{\partial X}
  {\partial x_t(\textbf{or} \ \partial x_b)}= x_l + x_r,
\end{equation}
\begin{equation}
  \frac{\partial X}
  {\partial x_l(\textbf{or} \ \partial x_r)}= x_t + x_b.
\end{equation}

To compute the partial derivative of $I$ w.r.t $x$, marked as $\nabla _xI$:
\begin{equation}
 \frac{\partial I}{\partial x_t(\textbf{or} \ \partial x_b)}=
  \begin{cases}
   I_w, & \text{if}\ x_t < \widetilde x_t(\textbf{or} \ x_b < \widetilde x_b)\\
   0, & \text{otherwise},
  \end{cases}
\end{equation}
\begin{equation}
 \frac{\partial I}{\partial x_l(\textbf{or} \ \partial x_r)}=
 \begin{cases}
  I_h, & \text{if}\ x_l < \widetilde x_l(\textbf{or} \ x_r < \widetilde x_r)\\
  0, & \text{otherwise}.
 \end{cases}
\end{equation}

\vfill\eject
Finally we can compute the gradient of localization loss $\mathcal{L}$ w.r.t.
$x$:
\begin{equation}
 \begin{split}
  \frac{\partial {\mathcal L}}{\partial x}
  & =
  \frac{I(\nabla _xX - \nabla _xI) - U\nabla _xI}{U^2IoU} \\
  & = \frac{1}{U}\nabla _xX - \frac{U+I}{UI}\nabla _xI.
 \end{split}
 \label{Eqn:gradient}
\end{equation}

From Eqn.~\ref{Eqn:gradient}, we can have a better understanding of the $IoU$
loss layer: the $\nabla _xX$ is the penalty for the predict bounding box, which
is in a positive proportion to the gradient of loss; and the $\nabla _xI$ is
the penalty for the intersection area, which is in a negative proportion to the
gradient of loss. So overall to minimize the $IoU$ loss, the
Eqn.~\ref{Eqn:gradient} favors the intersection area as large as possible while
the predicted box as small as possible. The limiting case is the intersection
area equals to the predicted box, meaning a perfect match.

\section{UnitBox Network} \label{sec:sec3}

\begin{figure}
 \centering
 \includegraphics[width=0.45\textwidth]{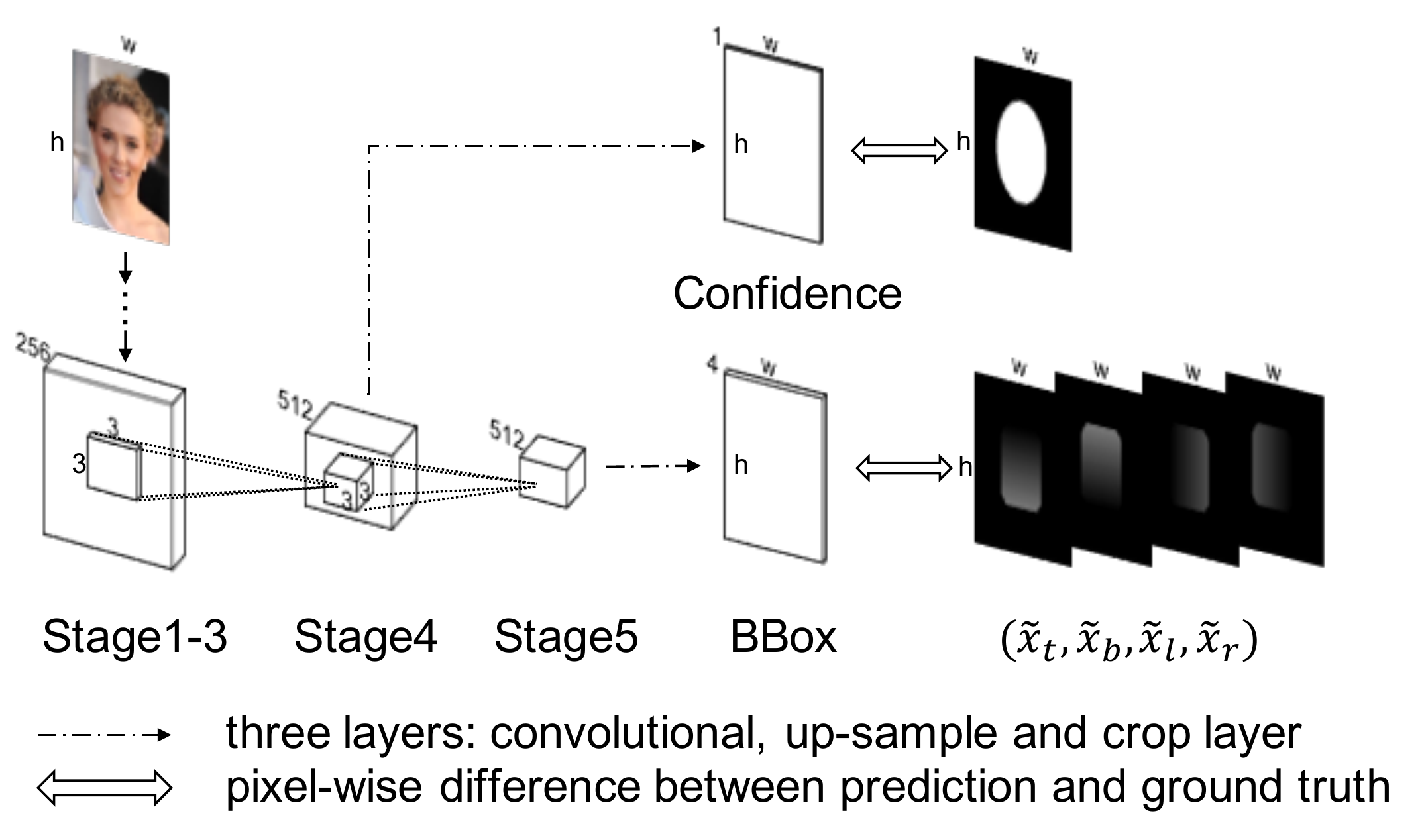}
  \caption{The Architecture of UnitBox Network.}
 \label{fig:network}
\end{figure}
Based on the $IoU$ loss layer, we propose a pixel-wise object detection
network, named UnitBox.  As illustrated in Figure~\ref{fig:network}, the
architecture of UnitBox is derived from VGG-16 model~\cite{cv:VGG}, in which we
remove the fully connected layers and add two branches of fully convolutional
layers to predict the pixel-wise bounding boxes and classification scores,
respectively.  In training, UnitBox is fed with three inputs in the same size:
the original image, the confidence heatmap inferring a pixel falls in a target
object (positive) or not (negative), and the bounding box heatmaps inferring
the ground truth boxes at all positive pixels.

To predict the confidence, three layers are added layer-by-layer at the end of
VGG stage-4: \textit{a convolutional layer} with stride $1$, kernel size
$512\times3\times3\times1$; \textit{an up-sample layer} which directly performs
linear interpolation to resize the feature map to original image size;
\textit{a crop layer} to align the feature map with the input image. After
that, we obtain a 1-channel feature map with the same size of input image, on
which we use the sigmoid cross-entropy loss to regress the generated confidence
heatmap; in the other branch, to predict the bounding box heatmaps we use the
similar three stacked layers at the end of VGG stage-5 with convolutional
kernel size 512 x 3 x 3 x 4. Additionally, we insert a ReLU layer to make
bounding box prediction non-negative. The predicted bounds are jointly
optimized with $IoU$ loss proposed in Section~\ref{sec:iou_loss}. The final
loss is calculated as the weighted average over the losses of the two branches.

Some explanations about the architecture design of UnitBox are listed as
follows: 1) in UnitBox, we concatenate the confidence branch at the end of VGG
stage-4 while the bounding box branch is inserted at the end of stage-5. The
reason is that to regress the bounding box as a unit, the bounding box branch
needs a larger receptive field than the confidence branch. And intuitively, the
bounding boxes of objects could be predicted from the confidence heatmap. In
this way, the bounding box branch could be regarded as a bottom-up strategy,
abstracting the bounding boxes from the confidence heatmap; 2) to keep UnitBox
efficient, we add as few extra layers as possible.  Compared to
DenseBox~\cite{cv:DenseBox} in which three convolutional layers are inserted
for bounding box prediction, the UnitBox only uses one convolutional layer. As
a result, the UnitBox could process more than 10 images per second, while
DenseBox needs several seconds to process one image; 3) though in
Figure~\ref{fig:network} the bounding box branch and the confidence branch
share some earlier layers, they could be trained separately with unshared
weights to further improve the effectiveness.

\begin{figure*}
 \centering
 \includegraphics[width=\textwidth]{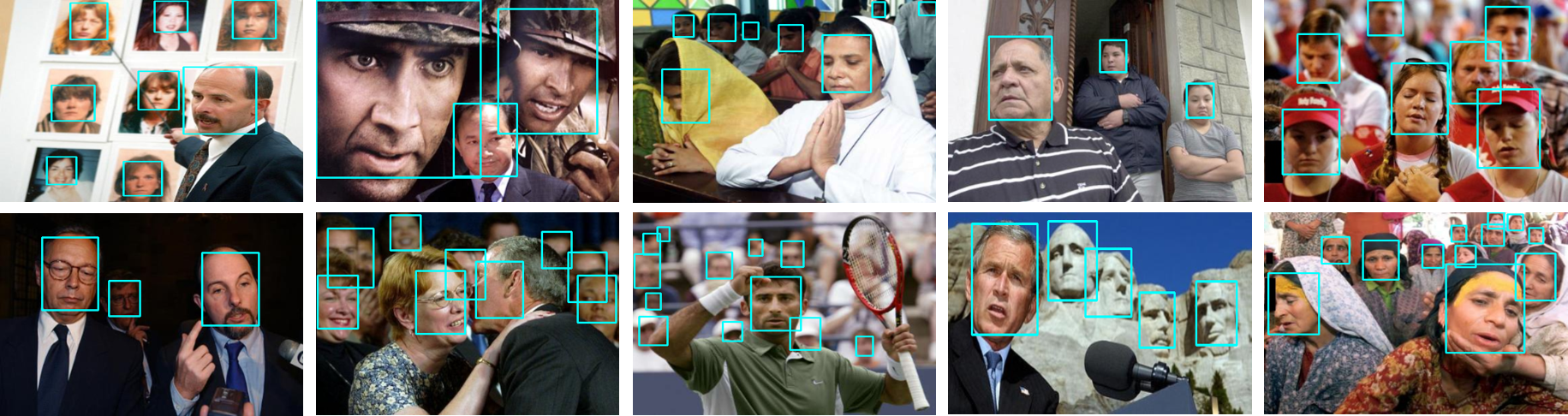}
 \caption{Examples of detection results of UnitBox on FDDB.}
 \label{fig:observation}
\end{figure*}

With the heatmaps of confidence and bounding box, we can now accurately
localize the objects.  Taking the face detection for example, to generate
bounding boxes of faces, firstly we fit the faces by ellipses on the
thresholded confidence heatmaps. Since the face ellipses are too coarse to
localize objects, we further select the center pixels of these coarse face
ellipses and extract the corresponding bounding boxes from these selected
pixels. Despite its simplicity, the localization strategy shows the ability to
provide bounding boxes of faces with high accuracy, as shown in
Figure~\ref{fig:observation}.

\section{Experiments}

In this section, we apply the proposed $IoU$ loss as well as the UnitBox on
face detection task, and report our experimental results on the FDDB
benchmark~\cite{data:FDDB}. The weights of UnitBox are initialized from a VGG-16
model pre-trained on ImageNet, and then fine-tuned on the public face dataset
WiderFace~\cite{data:wider_face}. We use mini-batch SGD in fine-tuning and set
the batch size to 10. Following the settings in~\cite{cv:DenseBox}, the
momentum and the weight decay factor are set to 0.9 and 0.0002, respectively.
The learning rate is set to $10^{-8}$ which is the maximum trainable value.  No
data augmentation is used during fine-tuning.

\subsection{Effectiveness of IoU Loss}

\begin{figure}[h]
 \subfigure[Convergence]
 {
 \includegraphics[width=0.22\textwidth]{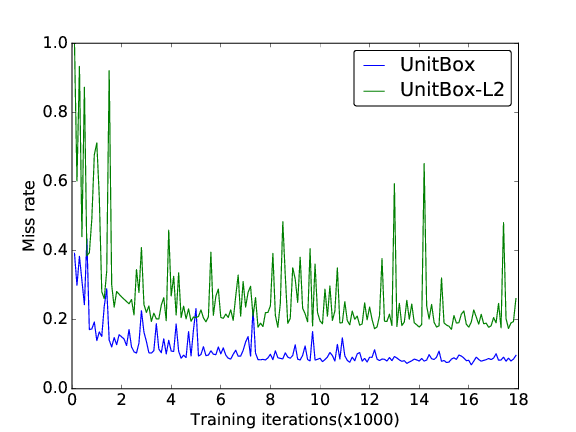}
 \label{fig:convergence}
 }
 \subfigure[ROC Curves]
 {
 \includegraphics[width=0.22\textwidth]{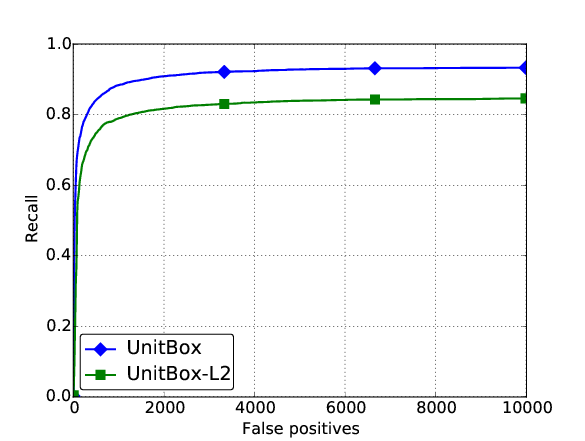}
 \label{fig:performance_vs}
 }
\caption{Comparison: $IoU$ vs. $\ell_2$.}
\end{figure}

First of all we study the effectiveness of the proposed $IoU$ loss. To train a
UnitBox with $\ell_2$ loss, we simply replace the $IoU$ loss layer with the
$\ell_2$ loss layer in Figure~\ref{fig:network}, and reduce the learning rate
to $10^{-13}$ (since $\ell_2$ loss is generally much larger, $10^{-13}$ is the
maximum trainable value), keeping the other parameters and network architecture
unchanged. Figure~\ref{fig:convergence} compares the convergences of the two
losses, in which the X-axis represents the number of iterations and the Y-axis
represents the detection miss rate.  As we can see, the model with $IoU$ loss
converges more quickly and steadily than the one with $\ell_2$ loss. Besides,
the UnitBox has a much lower miss rate than the UnitBox-$\ell_2$ throughout the
fine-tuning process.

In Figure~\ref{fig:performance_vs}, we pick the best models of UnitBox ($\sim$
16k iterations) and UnitBox-$\ell_2$ ($\sim$ 29k iterations), and compare their
ROC curves. Though with fewer iterations, the UnitBox with $IoU$ loss still
significantly outperforms the one with $\ell_2$ loss.

\begin{figure}[h]
 \includegraphics[width=0.45\textwidth]{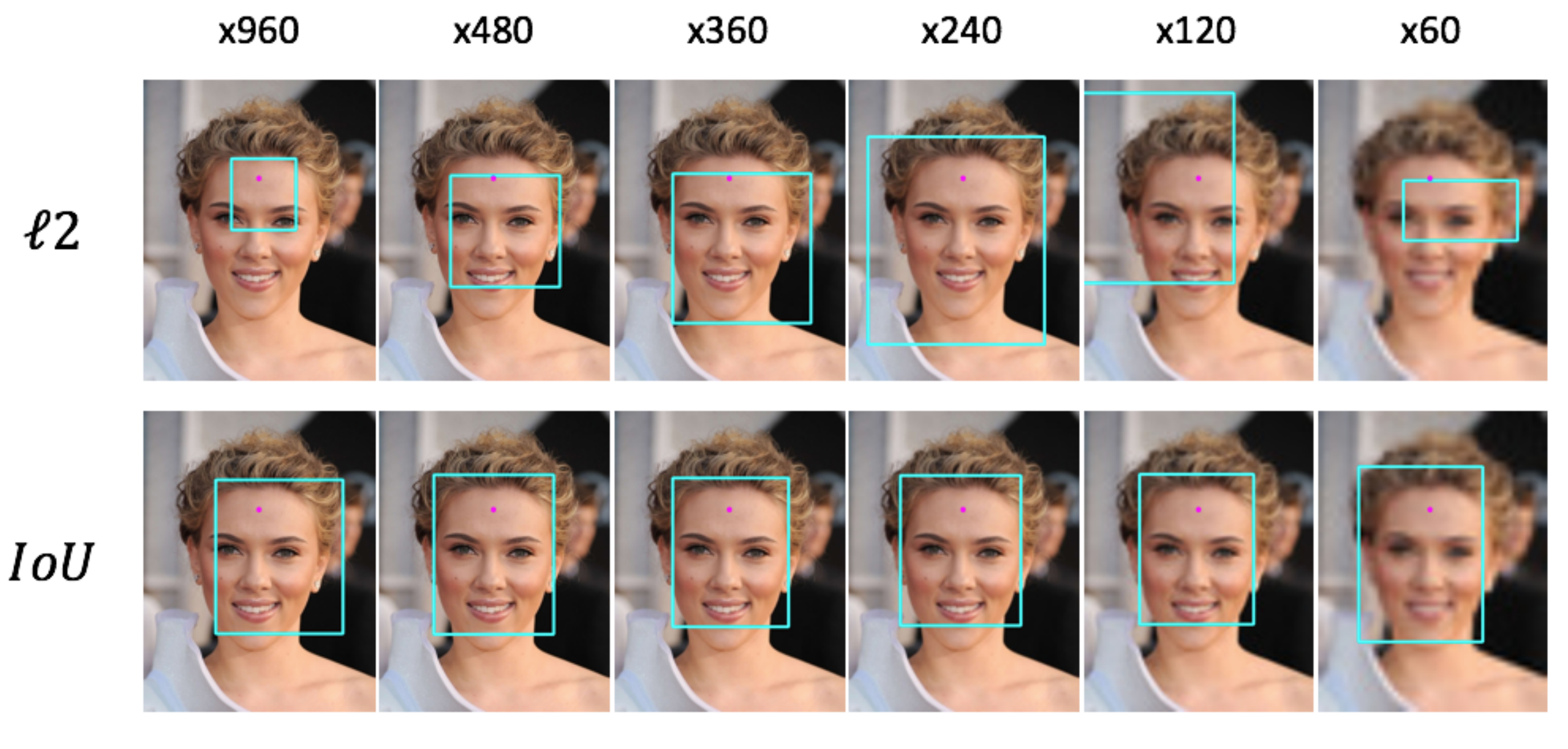}
 \caption{Compared to $\ell_2$ loss, the $IoU$ loss is much more robust to scale variations for bounding box prediction.}
 \label{fig:scale_invariant}
\end{figure}

Moreover, we study the robustness of $IoU$ loss and $\ell_2$ loss to the scale
variation.  As shown in Figure~\ref{fig:scale_invariant}, we resize the testing
images from 60 to 960 pixels, and apply UnitBox and UnitBox-$\ell_2$ on the
image pyramids. Given a pixel at the same position (denoted as the red dot),
the bounding boxes predicted at this pixel are drawn. From the result we can
see that 1) as discussed in Section ~\ref{ssec:ssec2.1}, the $\ell_2$ loss
could hardly handle the objects in varied scales while the $IoU$ loss works
well; 2) without joint optimization, the $\ell_2$ loss may regress one or two
bounds accurately, e.g., the up bound in this case, but could not provide
satisfied entire bounding box prediction; 3) in the x960 testing image, the
face size is even larger than the receptive fields of the neurons in UnitBox
(around 200 pixels). Surprisingly, the UnitBox can still give a reasonable
bounding box in the extreme cases while the UnitBox-$\ell_2$ totally fails.

\subsection{Performance of UnitBox}

\begin{figure}
 \includegraphics[width=0.45\textwidth]{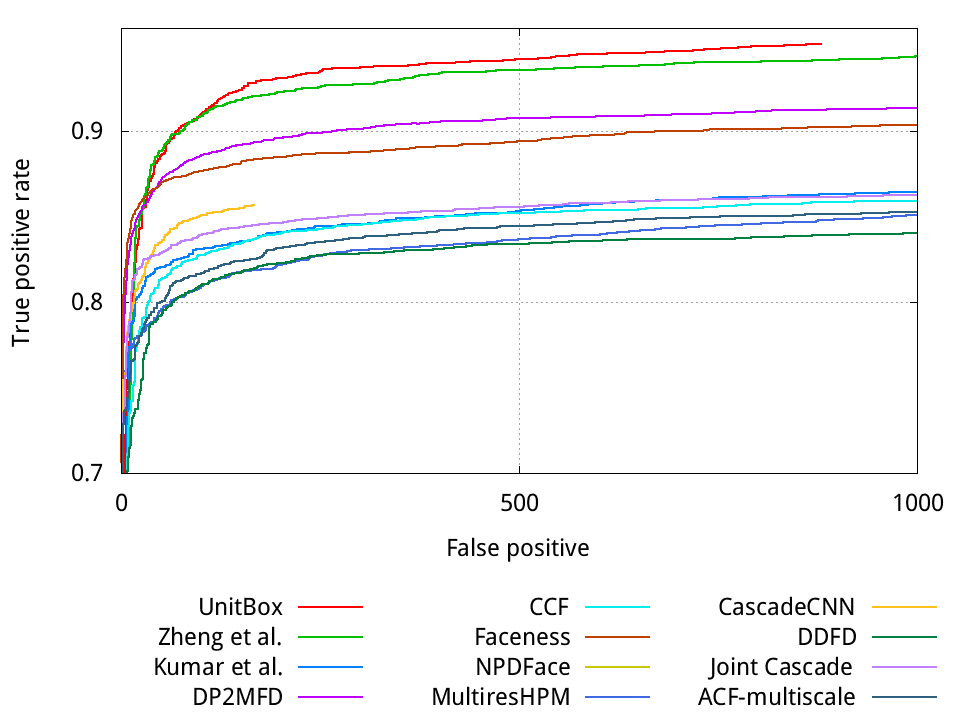}
 \caption{The Performance of UnitBox comparing with state-of-the-arts Methods on FDDB.}
 \label{fig:performance}
\end{figure}

To demonstrate the effectiveness of the proposed method, we compare the UnitBox
with the state-of-the-arts methods on FDDB. As illustrated in
Section~\ref{sec:sec3}, here we train an unshared UnitBox detector to further
improve the detection performance.  The ROC curves are shown in
Figure~\ref{fig:performance}. As a result, the proposed UnitBox has achieved
the best detection result on FDDB among all published methods.

Except that, the efficiency of UnitBox is also remarkable. Compared to the
DenseBox~\cite{cv:DenseBox} which needs seconds to process one image, the
UnitBox could run at about 12 fps on images in VGA size. The advantage in
efficiency makes UnitBox potential to be deployed in real-time detection
systems.

\section{Conclusions}
The paper presents a novel loss, i.e., the $IoU$ loss, for bounding box
prediction.  Compared to the $\ell_2$ loss used in previous work, the $IoU$
loss layer regresses the bounding box of an object candidate as a whole unit,
rather than four independent variables, leading to not only faster convergence
but also more accurate object localization. Based on the $IoU$ loss, we further
propose an advanced object detection network, i.e., the UnitBox, which is
applied on the face detection task and achieves the state-of-the-art
performance. We believe that the $IoU$ loss layer as well as the UnitBox will
be of great value to other object localization and detection tasks.

\vfill\eject
\bibliographystyle{abbrv}

\end{document}